\documentclass[letterpaper, 10 pt, conference]{ieeeconf}  %

\IEEEoverridecommandlockouts                              %

\usepackage{amssymb}
\usepackage{amsmath}
\usepackage{comment}
\usepackage{graphicx}
\usepackage{subcaption}

\usepackage{float}

\newcommand{\R}{\mathbb{R}}

\title{\LARGE \bf
Torque and velocity controllers to perform jumps with a humanoid robot: theory and implementation on the iCub robot
}

\author{Fabio Bergonti, Luca Fiorio and Daniele Pucci$^{1}$%
\thanks{$^{1}$Bergonti, Fiorio  and  Pucci  are  with  the  Fondazione  Istituto Italiano di Tecnologia, 16163 Genova, Italy (e-mail: {\tt\small name.surname@iit.it})} }
\begin{document}

\maketitle
\thispagestyle{empty}
\pagestyle{empty}

\begin{abstract}
Jumping can be an effective way of locomotion to overcome small terrain gaps or obstacles.
In this paper we propose two different approaches to perform jumps with a humanoid robot.
Specifically, starting from a pre-defined CoM trajectory we develop the theory for a velocity controller and for a torque controller based on an optimization technique for the evaluation of the joints input.
The controllers have been tested both in simulation and on the humanoid robot iCub.
In simulation the robot was able to jump using both controllers, while the real system jumped with the velocity controller only.
The results highlight the importance of controlling the centroidal angular momentum and they suggest that the joint performances, namely maximum power, of the legs and torso joints, and the low level control performances are fundamental to achieve acceptable results.
\end{abstract}

\section{INTRODUCTION}
The problem of locomotion for humanoid robots is an active research topic nowadays.
As humans, we have the tendency to mainly focus our attention on walking, but nature has shaped our legs for different locomotion approaches.
Among these different approaches, an interesting and challenging one is jumping.
A jump is mainly constituted by three different stages: launching, aerial phase and landing.
The aerial phase, in particular, requires a careful planning because of its uncontrollable nature, while the launching and landing phases involve high velocities and forces in short periods of time.

The problem of making a robot jumping is not new to the research community.
During the years, different researches have tackled the problem using different approaches and implemented the control algorithms on different platforms.
As an example, one of the first jumping robot is the planar one-legged hopping machine designed by Marc Raibert \cite{Raibert:1986:LRB:6152}.
The robot was constituted only by a body and a telescopic leg.
The leg was actuated by an air cylinder and at the end there was a padded foot.
The robot runs by hopping like a kangaroo.
In order to control the jump height the thrust of the pneumatic leg was controlled, while the posture of the main body was regulated to stabilize the system and adjust the forward speed.
Another interesting jumping pneumatic robot is Mowgli \cite{Niiyama2007}.
The robot design was inspired by the bio-mechanics of biological musculoskeletal structures.
Specifically the system is structured with a main body and two tampered legs with hip, knee, and ankle joints (3 Degrees of Freedom for each leg).
Its artificial musculoskeletal system consists of six McKibben pneumatic actuators including bi-articular configuration.
Thanks to the compliance, rapid contraction and high power/weight ratio of its muscles, Mowgli was able to perform jumps as high as $0.5 [m]$.

When it comes to humanoid robots the most notably results have been achieved with the robots Atlas and Asimo.
Atlas is a bipedal humanoid robot primarily developed by the American robotics company Boston Dynamics.
It is high $1.5 [m]$, weights $75 [kg]$ and exploits hydraulic actuators for the most powerful joints.
In a recent video \cite{AtlasBFLP}, the robot jumps over boxes and even performs a back-flip maneuver. In \cite{Dai2014WholebodyMP} Dai published an algorithm that makes Atlas jump in simulation. The approach uses a simple dynamics model and a full kinematic model.
Asimo, instead, is a humanoid robot developed by Honda in 2000. 
It is high $1.3 [m]$ and weights $54 [kg]$.
The robot has $34$ Degrees of Freedom (DoF) and uses only electric actuators. 
As shown in a popular video \cite{asimoVideo}, the robot is capable of performing small jumps using both legs or even using only one leg. 
The feet soles integrate a soft rubber layer, probably to dump the impact with the terrain.
QRIO is another robot that has been developed by Sony, it is also able to perform jump as described in \cite{Nagasaka}. Its algorithm is based on dynamics filter and requires a constant angular moment while the robot is in air.
In another interesting work \cite{Sakka2005}, the authors presents an approach based on feet/ground reaction forces to make the HRP-2 robot perform vertical jumps in simulations.
In this case, the feet reaction forces remain constant during the entire jumping phase and directly depend on the desired flight height.

While the results achieved by Boston Dynamics and Honda with their humanoid robots are impressive, little or none has been disclosed about how the robots are controlled.
Investigating the problem of making a humanoid robot jumping, and more in details implementing the control on a real platform, is a interesting task that calls for a fundamental improvement of the system at all levels.
Indeed, working on jumping with a humanoid robot is not only a control challenge, but also a design challenge.
The electronics, the linkages and the motors of the system are subject to high stresses.
During the launching phase the motors have to provide high torques at high velocity, namely high power.
The motor controllers have to provide sufficient current to power the motors and the control boards shall close the control loop at a high pace using accurate sensors.
Furthermore, during the landing phase, the linkages, bearings and transmissions are subject to peak forces and torques.

In this sense, the main contribution of the work is the presentation of two possible approaches to perform a controlled jump with a humanoid robot together with the implementation in simulation and on a real platform.
We propose a simple yet effective control strategy to jump by controlling the robot joints either in velocity or in torque, and we present the problems and limitations that we encountered on our platform.

The work is structured as follows. Section \ref{sec:background} introduces the mathematical formulation of the dynamics of mechanical systems commonly used in the whole-body control formulation together with a detailed analysis of the jump phases. Section \ref{sec:controller} describes the architecture of the proposed controllers and their respective key elements. Section \ref{sec:simExp} reports the results achieved both in simulation and with the real robot. Finally Section \ref{sec:conclusions} draws the conclusion and present the future work.
\section{BACKGROUND} \label{sec:background}
\subsection{Notation}
The following notation is used throughout the paper:
\begin{itemize}
    \item $s$ is the vector of joints positions
    \item $\tau$ is the vector of joints torques
    \item $\nu$ is the vector of base velocity and joints velocities
    \item $J_{c}$ is the Jacobian matrix of the Centre of Mass (CoM)
    \item $J_{lf}$ and $J_{rf}$ are the Jacobian matrix of the left and right foot respectively
    \item $J_M$ is the centroidal momentum matrix
    \item $ \dot{x}_{d_{c}}$ $x_{d_{c}}$ are the velocity and position of the CoM
\end{itemize}

\subsection{Modelling}
It is assumed that the robot is composed of $n+1$ rigid bodies, called links, and they are connected by $n$ joint with one degree of freedom each.
The robot has been modelled considering it is a \textit{free floating} system, meaning that it is not possible to define an \textit{a priori} constant pose for any link with respect to the inertial frame. 
As a consequence, the system possess \textit{n + 6} DoF. The configuration space of the robot is then described by a frame attached to one of the robot's link (usually called the \textit{base link}) and the joints position.
Indeed, the configuration space can be represented by a triplet $q = \left( I_{pB}, I_{RB}, s \right)$ where $\left( I_{pB}, I_{RB} \right)$  describes the origin and orientation of the \textit{base frame}, and $s$ denotes the joint angles.
The velocity of the multi-body system can be described as
$$
\nu = \left( I_{\dot{p}B}, I_{\omega B}, \dot{s} \right)
$$
where $I_{\omega B}$ is the angular velocity of the base expressed with respect to the inertial frame.

By assuming that the multi-body system is in contact with the environment through $n_c$ distinct contacts, and by relying on the Euler-Poincare formalism \cite{EulerPoicare} (Ch. 13.5) we can derive the following equations of motion for the multi-body system:
\begin{equation} \label{eq:multiBody}
M(q) \dot{\nu} + h(q,\nu) = B \tau + \sum_{k=1}^{n_c} J^T_{C_k} f_k
\end{equation}
where $M \in \R^{n+6 \times n+6}$ is the mass matrix, $h \in \R^{n+6}$ is the sum of the gravity and Coriolis term. $\tau$ are the internal actuation torques and $B$ is a selector matrix which depends on the available actuation. $f_k = {\left[ F^T_k , \mu^T_k \right]}^T \in \R^6$, with $F_k, \mu_k \in \R^3$ respectively the force and corresponding moment of the force, denotes an external wrench applied by the environment on the link of the $k$-th contact.
The Jacobian $J_{C_k} = J_{C_k} (q)$ is the map between the robot velocity $\nu$ and the linear and angular velocity
$
I_{vC_k} := ( I_{\dot{p} c_k} , I_{\omega C_k} )
$
of the frame $C_k$, i.e.
$
I_{c C_k} = J_{C_k} (q) \nu
$.

\subsection{Locked angular velocity} \label{sub:lockedAngular}
The \textit{locked angular velocity} can be defined as the angular velocity a multi-body system would have if all the joints were locked instantaneously, namely making the multi-body system a single \textit{equivalent rigid body}.
Having a low or zero angular velocity is fundamental to have a smoother transition between the aerial and landing phases and subsequent balancing.
The locked angular velocity is strictly related to the centroidal angular momentum $H_{\omega}$.
\begin{equation}
H =
\begin{bmatrix} H_l \\ H_{\omega} \end{bmatrix} =
\begin{bmatrix} J_M^{l} \\ J_M^{\omega} \end{bmatrix} \nu =
J_M \nu
\end{equation}
Indeed, for the conservation of the centroidal momentum, if $H_{\omega}$ is equal to zero the locked angular velocity will be equal to zero, which implies no rotation of the \textit{equivalent rigid body} during the aerial phase.

\subsection{Jump phases and take-off velocity} \label{sec:jumpPhases}
In general a jump is characterized by the following phases:
\begin{enumerate}
\item Launching phase
\item Aerial phase
\item Landing phase
\end{enumerate}
The launching phase is the most crucial as it allows to meet the jump requirements (desired height, flight and landing stability).
During the launching phase, the upward velocity of the CoM is increased by exploiting the contacts with the environment.
Humans, as an example, start from an upright standing position, make a preliminary downward movement by flexing the knees and hips, and then propel upwards.
The way this phase is performed influences the muscle "pre-stretch".
Experiments have demonstrated that pre-stretch enhances the force production and work output of the muscles in the subsequent movement \cite{Linthorne2001}.
On robots with stiff actuators is quite different, the joints can not be "pre-stretched", and generally can produce maximum torques from any static configuration.
Because of this characteristics, differently from humans, the best initial configuration is not the upright standing position, but rather a squat position.
In a squat jump, the jumper starts from a lowered stationary position, then vigorously extends the knees and the hips to reach the desired upward take-off velocity.

By assuming that there are no dissipation of energy, namely the effect of the air resistance is negligible, we can easily compute the desired take-off speed as follows:
\begin{tabbing}
\hspace{3mm} \= $\Pi$ \hspace{3mm} \= Total Energy (Kinetic + Potential Energy)\\
\> $h_{to}$ \> Take-off height. It occurs at $t=t_{to}$\\
\> $h_{p}$ \> Maximum height. It occurs at $t=t_{p}$\\
\end{tabbing}
\noindent
When $t=t_{to}$ the total energy is 
$\Pi_{to}=\frac{1}{2}mv^2_{to}+mgh_{to}$.
Instead when $t=t_{p}$ the total energy is
$\Pi_{p}=mgh_{p}$.
Using the hypothesis of no dissipation of energy $\Pi_{p}=\Pi_{to}$ is it possible to compute the vertical take-off speed as:
$$v_{to}=\sqrt{2 g \left( h_{p}-h_{to} \right)}$$

As soon as the system looses contact with the environment, the aerial phase starts.
During this phase it is crucial to correctly prepare for the landing phase, because of the high ground reaction forces and sharp impacts involved.
Specifically, during landing, ground reaction forces can reach levels of an order of magnitude higher than the gravitational/weight forces.

\section{CONTROL FRAMEWORK} \label{sec:controller}

In this section we describe the control framework used in our work.
The controller is structured in two different modules.
As described in \ref{sub:Center of Mass Trajectory}, the first module computes the trajectory of the CoM to reach a specific take-off velocity.
Subsequently, for the launching phase, we rely on an instantaneous optimization technique to solve for the robot control inputs.
The optimization has been implemented in two different ways.
The first approach, described in \ref{sub:velocityMode}, solves for the joint velocities, while the second approach, described in \ref{sub:torqueMode} is torque based.
The aerial and landing phases instead, are managed using position control.

\subsection{Vertical Center of Mass Trajectory} \label{sub:Center of Mass Trajectory}
The CoM trajectory profile is obtained from a pre-designed curve (like the one presented in \cite{Linthorne2001}).
We designed a normalized jump trajectory with a final take-off speed equal to $1 [m/s]$ that is reached in a time equal to $1 [s]$.
The time parameter, in particular, can be computed in two different ways.
One possibility is to impose the final acceleration at the take-off equal to $-9.81 [m/s^2]$, this could be useful if the goal is to achieve continuity in the CoM acceleration.
Another possibility is to set the value of the displacement of the CoM during the launching phase. 
Among the two possible implementations, we selected the latter in order to avoid knee singular configurations.

In Fig.\ref{fig:CoMSpeed} is depicted an example of the the output of this procedure: the desired vertical trajectory of the CoM in terms of acceleration and velocity.
\begin{figure}
        \begin{subfigure}[b]{0.49\columnwidth}
            \centering
            \includegraphics[width=\textwidth]{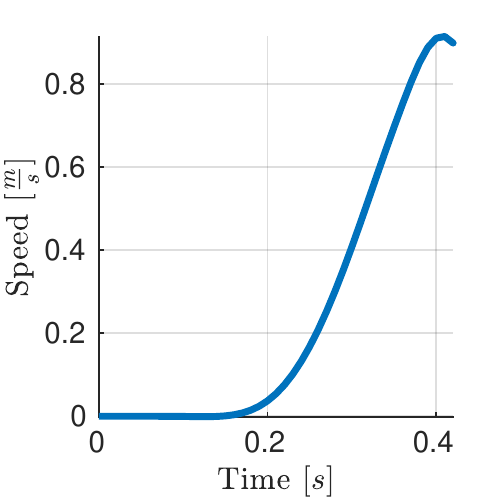}
            \caption[Network2]%
            {{Vertical CoM Speed}}    
            \label{fig:mean and std of net14}
        \end{subfigure}
        \hfill
        \begin{subfigure}[b]{0.49\columnwidth}  
            \centering 
            \includegraphics[width=\textwidth]{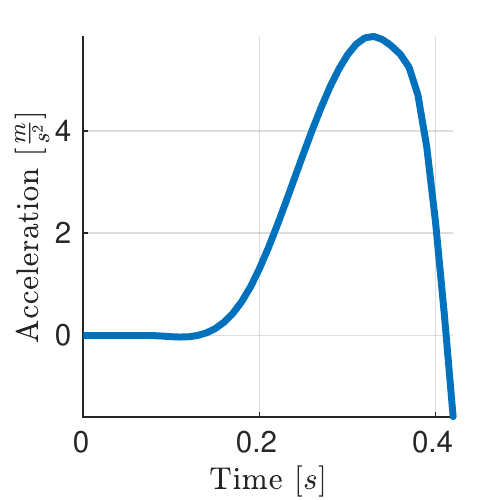}
            \caption[]%
            {{Vertical CoM Acceleration}}    
            \label{fig:mean and std of net24}
        \end{subfigure}
        \caption[ Velocity Acceleration profiles ]
        {\small A desired vertical profile of the CoM obtained setting as initial parameters a take off speed $0.89 [\frac{m}{s}]$ (a desired jumping height of $4 [cm]$), and a variation of the position of the CoM in the launching phase equal to $11 [cm]$.} 
        \label{fig:CoMSpeed}
\end{figure}

\subsection{Launching Phase} \label{sub:Optimization problem}
The CoM trajectory is defined in the Cartesian space.
The problem of mapping the Cartesian task to the joint space has been solved by implementing an instantaneous optimization algorithm.
More in details, we defined a cost function as a linearly constrained quadratic optimization problem that has been solved by relying on a Quadratic Programming (QP) approach.
This optimization problem can be formulated as:
\begin{IEEEeqnarray*}{LL}
	\min_{\nu}    & \frac{1}{2} \nu^T H \nu + \nu^T g  \\
	\text{s.t.}: & l_b \le A \nu \le u_b  
\end{IEEEeqnarray*}
Where $H$ is the Hessian matrix, $g$ is the gradient vector, $A$ is the constraint matrix, $l_b$ is the lower constraint vector and  $u_b$ is the upper constraint vector.
In our optimization problem we have added three types of constraints:
\begin{itemize}
    \item Task Constraint
    \item Behaviour Constraint
    \item System Constraint
\end{itemize}
The task constraints guarantee that the CoM follows the desired trajectory in order to reach the desired take-off velocity.
The behaviour constraints are needed to maintain balance and to complete the task without falling.
The system constraints guarantee the feasibility of the solution because force the solver to respect the speed and Range of Motion\footnote{The joint position is simply obtained integrating the joint velocity using forward Euler method.
byIn particular $s = s_ {m} + \dot{s} T_ {s}$ where $s_m$ is the measured joint position and $T_{s}$ is iteration time.} (RoM) joints limits.

\subsection{Aerial and Landing Phases} \label{sub: Aerial and Landing Phases}
As soon as there is no more contact with the ground the robot enters in the aerial phase.
During this phase the robot is switched to position control mode and the joints are moved in a short period of time to a specific landing configuration.
This configuration has been chosen in order to increase the maximum height reached by the feet (i.e. the distance between feet and center of mass is decreased) and to be stable once the robot lands on the floor.
The joint movements are computed using a dedicated minimum jerk trajectory generator \cite{Pattacini2010}.
Finally, during the landing phase the robot configuration is not changed. 
Technically this approach is not the best solution, but considering that the jump height is limited, the impact forces and joint torques are not a real issue.

\subsection{Velocity approach} \label{sub:velocityMode}
The optimization problem for the velocity approach has been structured considering that a humanoid robot is a free-floating system.
More in details, the vector of unknowns $\nu \in \mathbb{R}^{6+n}$ includes the joints velocities $\dot{s}$ $\in \mathbb{R}^n$ together with the base velocity $V_b \in \mathbb{R}^6$.
\begin{equation*}
    \nu =
    \begin{bmatrix}
        V_b \\
        \dot{s}
    \end{bmatrix}
\end{equation*}
The last $n$ rows of the vector $\nu$ represent the control inputs sent to the robot.\\
The optimization problem has been formulated as follows:
\begin{IEEEeqnarray}{LL}
	\min_{\nu}    & \frac{1}{2} \| \dot{s} - \dot{s}^* \|^2_{\Lambda} + \frac{1}{2} \| \dot{s} - \dot{s}_{k-1} \|^2_{\Delta} \IEEEyessubnumber \label{eq: Cost Function} \\ 
	 \text{s.t.}: & J_{c} \nu = \dot{x}_{d_{c}} - K \left( x-x_{d _{c}} \right)  \IEEEyessubnumber \label{eq: CoM Speed}\\
	& J_{lf} \nu = 0 \, , \; J_{rf} \nu = 0 \IEEEyessubnumber \label{eq: lf Speed} \\
	& J_M^{\omega} \nu = - K \int_ 0 ^t  H_ {\omega} d \tau \IEEEyessubnumber \label{eq: AngularMomentumConstraint}\\
	& \dot{s}^- \le \dot{s} \le  \dot{s}^+  \, , \; s^- \le s \le s^+
	\IEEEyessubnumber \label{eq: Joint Velocity}
\end{IEEEeqnarray}
The cost function (Eq.\ref{eq: Cost Function}) is composed of two terms.
The first term is the postural task, where the fictitious velocity is defined as $\dot{s}^* = - K_p \left( s-s_d \right)$ with $s_d$ as the postural position. In our case, we have selected $s_d$ equal to the initial joint position.
The second term of the cost function has the goal of minimizing the difference between $\dot{s}$ and $\dot{s}_{k-1}$, namely the joints velocities evaluated in the previous iteration. 
This term has been added in order to favour small variation of the control input and guarantee continuity.
Eventually, two definite positive matrices, $\Delta$ and $\Lambda$, are used to weight the two terms. 
The matrices can also be used to favour movement or continuity of some joints with respect to the others by acting on the respective diagonal elements.

Concerning the constraints we implemented the task constraint \ref{eq: CoM Speed} using the Jacobian matrix of the CoM because it relates linearly the joint speed $\dot{s}$ with the Cartesian CoM velocity $\dot{x}_{d_{c}}$.
We also added an extra term to close the position feedback loop by using the estimated position of the CoM $x$, evaluated assuming that the feet are in contact with the ground and they don't slide nor rotate.

The behaviour constraints are implemented in Eq.\ref{eq: lf Speed} and \ref{eq: AngularMomentumConstraint}.
More in details, Eq.\ref{eq: lf Speed} are two holonomic constraints needed to impose feet speed equal to zero, otherwise the robot could move them in the launching phase.  
Equation \ref{eq: AngularMomentumConstraint} instead, is used to reduce the robot rotation during the aerial phase (see \ref{sub:lockedAngular} for more details).
The constraint is structured like a first order dynamical system $\dot{x}= -K x$. 
If K is positive, $x$ will tend towards $0$.
\begin{gather}
    H_ {\omega} = - K \int_ 0 ^t  H_ {\omega} d \tau \label{eq: Momentum 0} \\
    \dot{H}_{\omega} = - K  H_{\omega} \label{eq: Momentum 1}
\end{gather}
where $\int_ 0 ^t  H_ {\omega} d \tau$ is evaluated as the numerical integration of the measured centroidal angular momentum.

Finally, the system constraints are implemented in Eq.\ref{eq: Joint Velocity}.%

\subsection{Torque approach} \label{sub:torqueMode}
In case of torque control, the optimization problem is formulated in a different way.
\begin{IEEEeqnarray}{LL}
	\min_{u}    & \frac{1}{2} \| \ddot{s} - \ddot{s}^* \|^2_{\Lambda} + \frac{1}{2} \| \tau - \tau_{k-1} \|^2_{\Delta} \hspace{5mm} \textrm{with } u = \begin{bmatrix} \tau \\ f \end{bmatrix} \IEEEyessubnumber \label{eq: Cost Function Torque} \\ 
	 \text{s.t.}: & J_{c} \dot{\nu} + \dot{J}_{c} \nu = \ddot{x}_{d_{c}} -
	 \begin{bmatrix} K_p & 0 \\ 0 & K_d \end{bmatrix}
	 \begin{bmatrix} x-x_{d _{c}} \\ \dot{x}-\dot{x}_{d _{c}} \end{bmatrix}
	 \IEEEyessubnumber \label{eq: CoM Acceleration}\\
	& J_{lf} \dot{\nu} + \dot{J}_{lf} \nu = 0   \, , \; J_{rf} \dot{\nu} + \dot{J}_{rf} \nu= 0 \IEEEyessubnumber \label{eq: lf Acceleration} \\
	& A f = - K H_ {\omega} \IEEEyessubnumber \label{eq: AngularMomentumConstraint Torque}\\
	& \dot{s}^- \le \dot{s} \le  \dot{s}^+ \, , \;
	s^- \le s \le s^+ \, , \;
	\tau^- \le \tau \le \tau^+ \IEEEyessubnumber \label{eq: Joint Velocity Torque}\\
	& C u \le b \IEEEyessubnumber \label{eq: cu b}
\end{IEEEeqnarray}
Where the term $\dot{\nu}$ can be computed by extending the multi-body formulation of Eq.\ref{eq:multiBody} including the vector $f = \begin{bmatrix} f_{lf} \\ f_{rf} \end{bmatrix}$ containing the ground reaction forces for both feet.
\begin{gather*}
    M(q) \dot{\nu} + h(q,\nu) = B \tau + J_ {lf}^T f_{lf} + J_ {rf}^T f_{rf} \\
    \dot{\nu} = M(q)^{-1} \left( B \tau + J_{f} f - h(q,\nu) \right) \; \,
    \textrm{with } J_{f} = \begin{bmatrix} J_ {lf}^T & J_ {rf}^T \end{bmatrix}
\end{gather*}
Similarly to the velocity approach, the torque cost function \ref{eq: Cost Function Torque} is constituted by two terms.
In this case, the postural task is implemented through the fictitious acceleration $\ddot{s}^* = K_d \dot{s} - K_p \left( s-s_d \right)$, while continuity is guaranteed again by minimizing the difference with the acceleration computed in the previous iteration.

The task constraint (eq \ref{eq: CoM Acceleration}), and the first two behaviour constraints (eq \ref{eq: lf Acceleration}) are expressed in this case using the joints accelerations.
The constraint on the angular momentum is written in another form. 
Specifically, considering that the time derivative of the momentum is equal to the external forces (the ground reaction forces $f$), we evaluate $f$ such that it constraints the angular momentum to behave like the first order system described in Eq.\ref{eq: Momentum 1}.
For this purpose, $A$ is a projection matrix used to combine the effect of the linear ground reaction forces with the ground reaction moments.
The additional behavior constraint (\ref{eq: cu b}) ensure the satisfaction of friction cones, normal contact surface forces, and Center of Pressure (CoP) constraints (i.e. the CoP must lay within the foot support polygon).

For what concerns system constraints, the joint speed and position limits are still present.
The torque limit has been added to take into account the limitations of the actuators.
\section{SIMULATIONS AND EXPERIMENTS} \label{sec:simExp}
We tested both the velocity and the torque approaches in simulation and on the real robot.
The robot chosen for the experiments is the iCub robot.
iCub is a humanoid robot with physical and cognitive abilities similar to the ones of a 5-year-old baby.
It's $1.04 [m]$ tall and it weights approximately $30 [kg]$ \cite{Natale2017}.
It's constituted by $53$ DoF, but only a total of $23$ are used for the jumping task, i.e. we do not consider those located in the neck, eyes and in the hands.
Each leg has $6$ DoF: $3$ in the hip pitch, $1$ in the knee and $2$ in the ankle \cite{Parmiggiani2012}.
The waist has other additional $3$ DoF.

In order to be able to perform the jump experiments on the real platform we had to improve the low level control and modify the robot joints.
More in details the following improvements have been adopted:
\begin{itemize}
    \item \textbf{Shifted the ankle RoM}: the ankle design has been modified to shift of $15 [deg]$ the \textit{ankle pitch} RoM in order to allow the robot to crouch down more.
    \item \textbf{Increased the low level current and voltage limits}: all the legs and torso current and voltage limits have been increased to the maximum value.
\end{itemize}
From the controller implementation standpoint, the QP problem has been solved using qpOASES \cite{Ferreau2014}, while the drivers have been developed in Matlab Simulink exploiting the Whole-Body Toolbox \cite{RomanoWBI17Journal}.
The jumping controller input parameters have been set to a desired jump height of $4 [cm]$ (desired take-off velocity $0.89 [\frac{m}{s}]$) with a displacement of the CoM during the launching phase of $11 [cm]$.
With this normalization the take-off occurs at $0.43 [s]$.

\subsection{Simulation results}
\begin{figure*}
        \begin{subfigure}[b]{0.49\columnwidth}
            \centering
            \includegraphics[width=\textwidth]{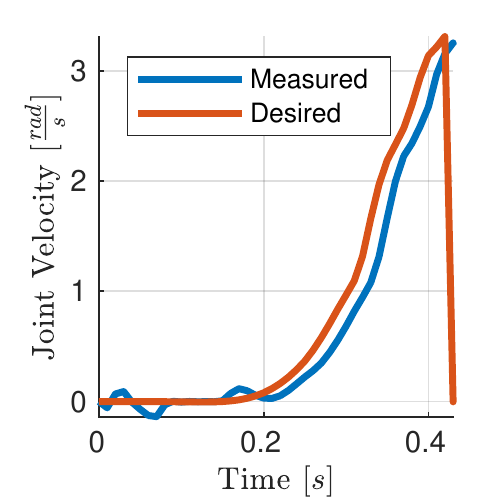}
            \caption[Network2]%
            {{Left Ankle Pitch}}    
            \label{fig:mean and std of net14}
        \end{subfigure}
        \hfill
        \begin{subfigure}[b]{0.49\columnwidth}  
            \centering 
            \includegraphics[width=\textwidth]{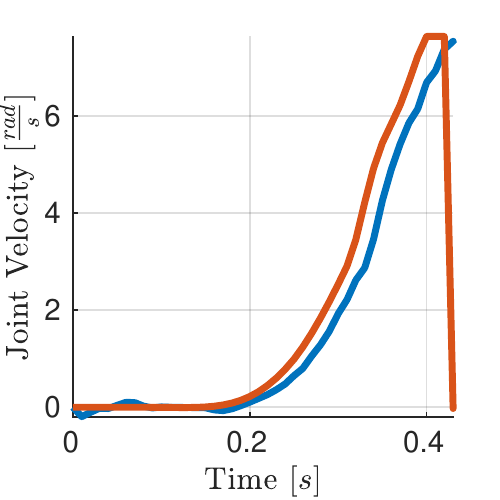}
            \caption[]%
            {{Left Knee}}    
            \label{fig:mean and std of net24}
        \end{subfigure}
        \begin{subfigure}[b]{0.49\columnwidth}   
            \centering 
            \includegraphics[width=\textwidth]{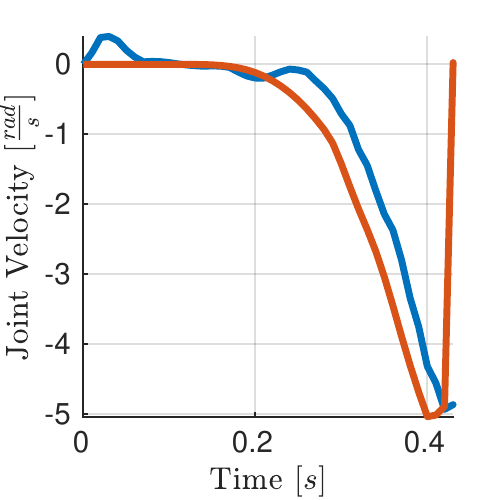}
            \caption[]%
            {{Left Hip Pitch}}    
            \label{fig:mean and std of net34}
        \end{subfigure}
        \quad
        \begin{subfigure}[b]{0.49\columnwidth}   
            \centering 
            \includegraphics[width=\textwidth]{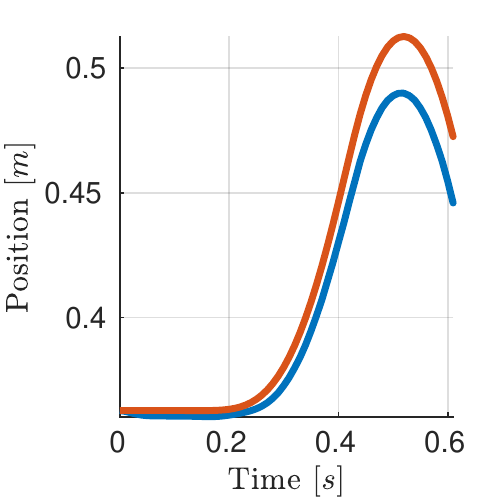}
            \caption[]%
            {{Z Position CoM}}    
            \label{fig:mean and std of net44}
        \end{subfigure}
        \caption[ Simulation in Velocity ]
        {\small The results of the simulation using the velocity approach. The take-off occurs at $0.43s$.} 
        \label{simVelocity}
\end{figure*}

The simulations were performed using the open source simulator Gazebo with a maximum step size of $0.0001 [s]$.
The results of the simulation with the velocity approach are depicted in Fig.\ref{simVelocity}.
The CoM height increases of $3.5 [cm]$ during the flight phase, while the maximum height reached by the feet soles is $3.7 [cm]$.

\begin{figure*}
        \begin{subfigure}[b]{0.49\columnwidth}
            \centering
            \includegraphics[width=\textwidth]{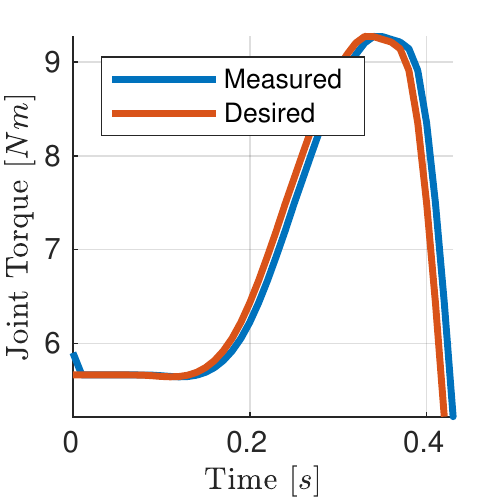}
            \caption[Network2]%
            {{Left Ankle Pitch}}    
            \label{fig:mean and std of net14}
        \end{subfigure}
        \hfill
        \begin{subfigure}[b]{0.49\columnwidth}  
            \centering 
            \includegraphics[width=\textwidth]{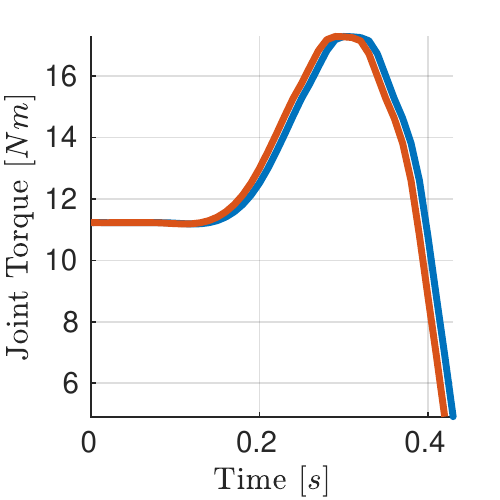}
            \caption[]%
            {{Left Knee}}    
            \label{fig:mean and std of net24}
        \end{subfigure}
        \begin{subfigure}[b]{0.49\columnwidth}   
            \centering 
            \includegraphics[width=\textwidth]{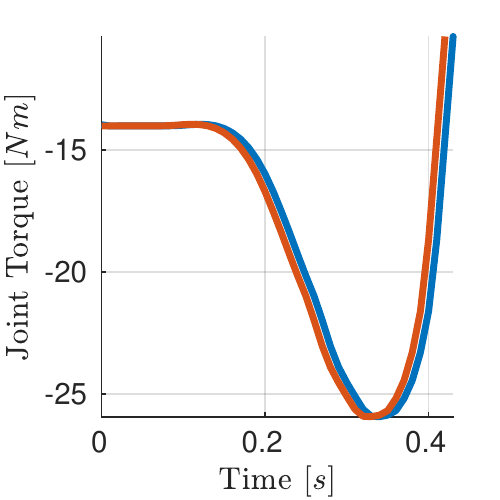}
            \caption[]%
            {{Left Hip Pitch}}    
            \label{fig:mean and std of net34}
        \end{subfigure}
        \quad
        \begin{subfigure}[b]{0.49\columnwidth}   
            \centering 
            \includegraphics[width=\textwidth]{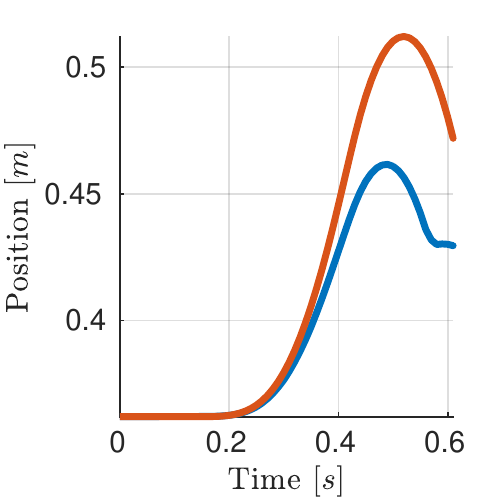}
            \caption[]%
            {{Z Position CoM}}    
            \label{fig:mean and std of net44}
        \end{subfigure}
        \caption[ Simulation in Velocity ]
        {\small The results of the simulation using the torque approach. The take-off occurs at $0.43s$} 
        \label{simTorque}
\end{figure*}

The results of the simulation with the torque approach are depicted in Fig.\ref{simTorque}.
The CoM height increases of $1.6[cm]$ during the flight phase, while the maximum height reached by the feet soles is $2.3 [cm]$.

The \textit{Measured position} defined in Fig.\ref{simVelocity} and Fig.\ref{simTorque} is the effective position of the CoM estimated by a custom Gazebo plugin during the entire simulation.

\subsection{Experimental results}
\begin{figure*}
        \begin{subfigure}[b]{0.49\columnwidth}
            \centering
            \includegraphics[width=\textwidth]{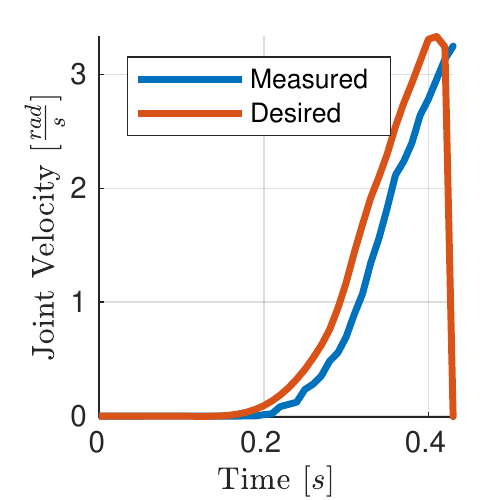}
            \caption[Network2]%
            {{Left Ankle Pitch}}    
            \label{fig:mean and std of net14}
        \end{subfigure}
        \hfill
        \begin{subfigure}[b]{0.49\columnwidth}  
            \centering 
            \includegraphics[width=\textwidth]{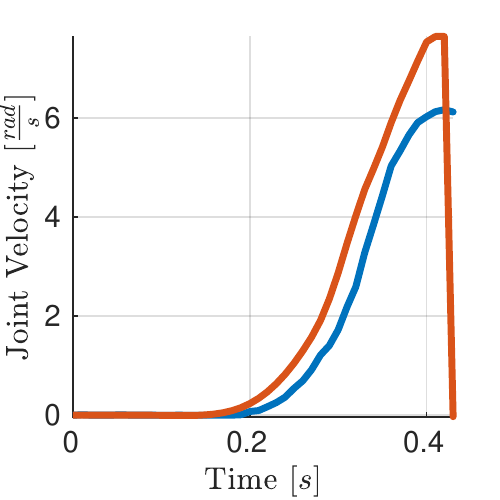}
            \caption[]%
            {{Left Knee}}    
            \label{fig:mean and std of net24}
        \end{subfigure}
        \begin{subfigure}[b]{0.49\columnwidth}   
            \centering 
            \includegraphics[width=\textwidth]{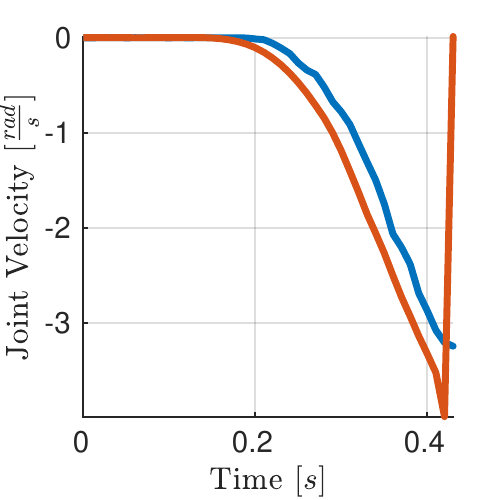}
            \caption[]%
            {{Left Hip Pitch}}    
            \label{fig:mean and std of net34}
        \end{subfigure}
        \quad
        \begin{subfigure}[b]{0.49\columnwidth}   
            \centering 
            \includegraphics[width=\textwidth]{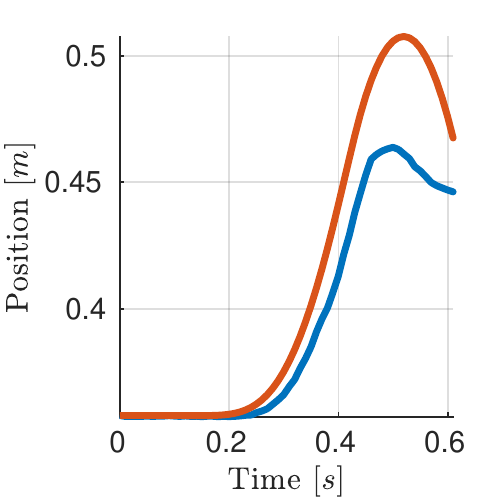}
            \caption[]%
            {{Z Position CoM}}    
            \label{fig:mean and std of net44}
        \end{subfigure}
        \caption[ Real robot in Velocity ]
        {\small The results of the experimental tests on iCub using the velocity approach. The take-off occurs at $0.43s$} 
        \label{realVelocity}
\end{figure*}

The results of the experimental test with the velocity approach are depicted in Fig.\ref{realVelocity}. 
The \textit{Measured position} of the CoM is estimated assuming that the feet are in contact with the ground.
As a consequence, after the take-off this estimation is no more reliable.
The measured take-off speed permits us to have an estimation of the CoM height increase during the flight phase, its value is $3 [cm]$, instead the feet soles height is higher thanks to the feet retraction during the flight phase. 

For what concerns the torque control, even if the results obtained from the simulations were satisfactory, with the real robot we couldn't complete a jump.
More in details, we encountered the following problems, that will be addressed in the future work:
\begin{itemize}
    \item \textbf{Computational time}: the Simulink drivers were not able to run real time due to the heavy computations of the QP.
    \item \textbf{Low level tuning}: the low level torque parameters were tuned for less dynamic tasks.
    \item \textbf{Errors in the estimated torques}: the measurements coming from the force-torque sensors are subject to drift and errors that influence the low level torque feedback.
\end{itemize}

\begin{figure*}
        \begin{subfigure}[b]{0.4\columnwidth}
            \centering
            \includegraphics[width=\textwidth]{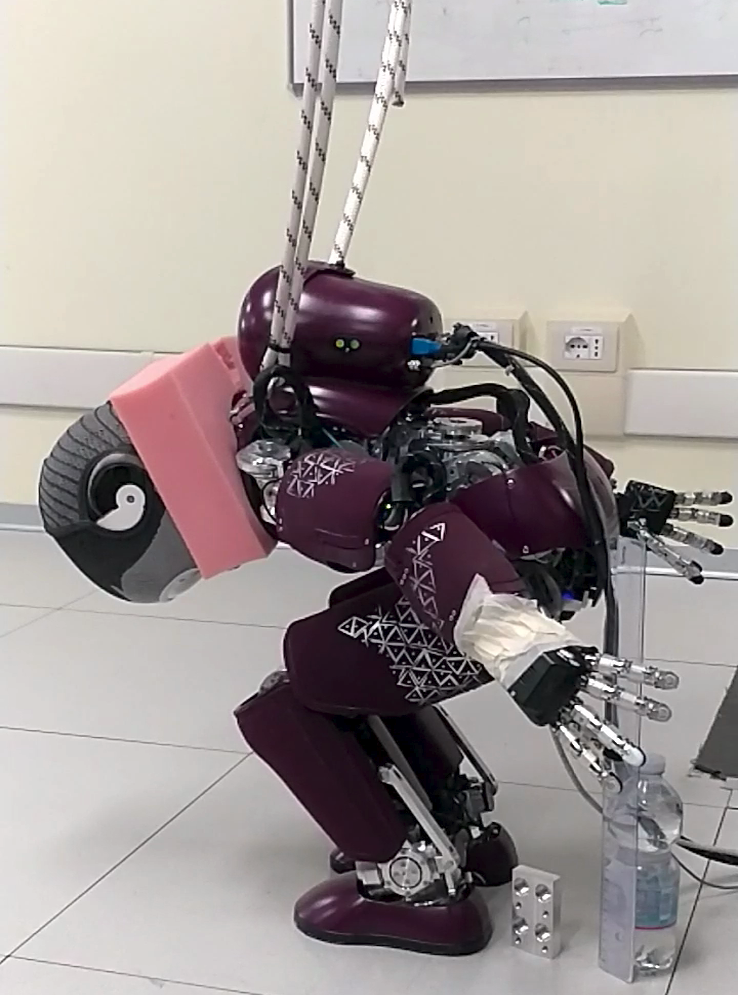}
            \caption[Network2]%
            {{Initial Configuration}}    
            \label{fig:mean and std of net14}
        \end{subfigure}
        \hfill
        \begin{subfigure}[b]{0.4\columnwidth}  
            \centering 
            \includegraphics[width=\textwidth]{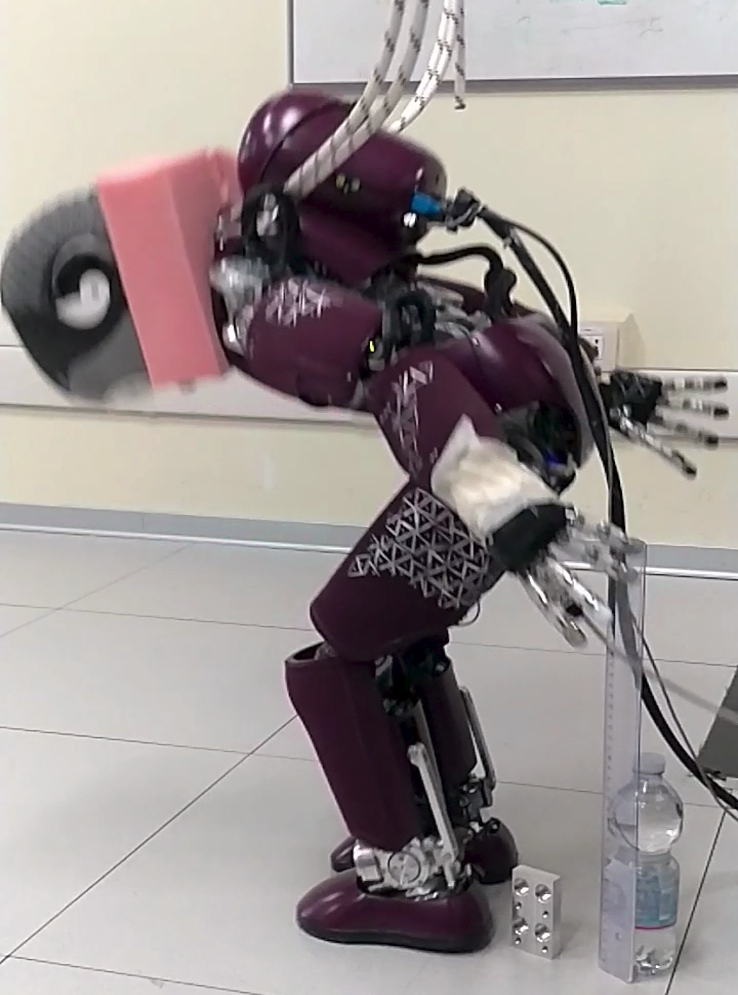}
            \caption[]%
            {{Take-off}}    
            \label{fig:mean and std of net24}
        \end{subfigure}
        \begin{subfigure}[b]{0.4\columnwidth}  
            \centering 
            \includegraphics[width=\textwidth]{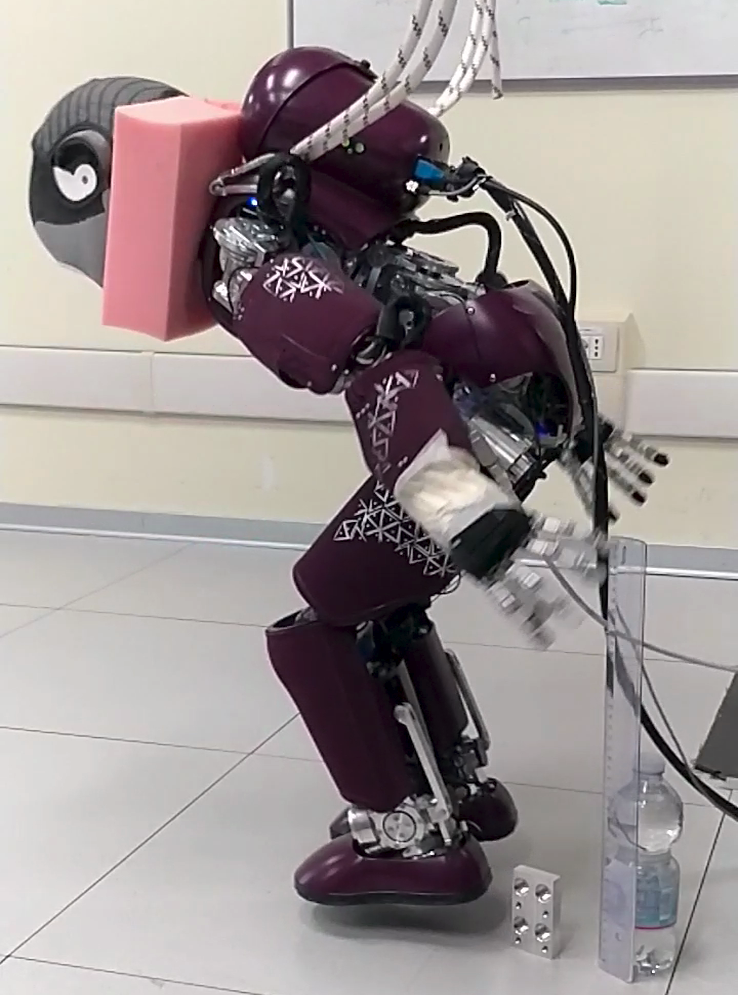}
            \caption[]%
            {{Maximum Height}}    
            \label{fig:mean and std of net24}
        \end{subfigure}
        \begin{subfigure}[b]{0.4\columnwidth}  
            \centering 
            \includegraphics[width=\textwidth]{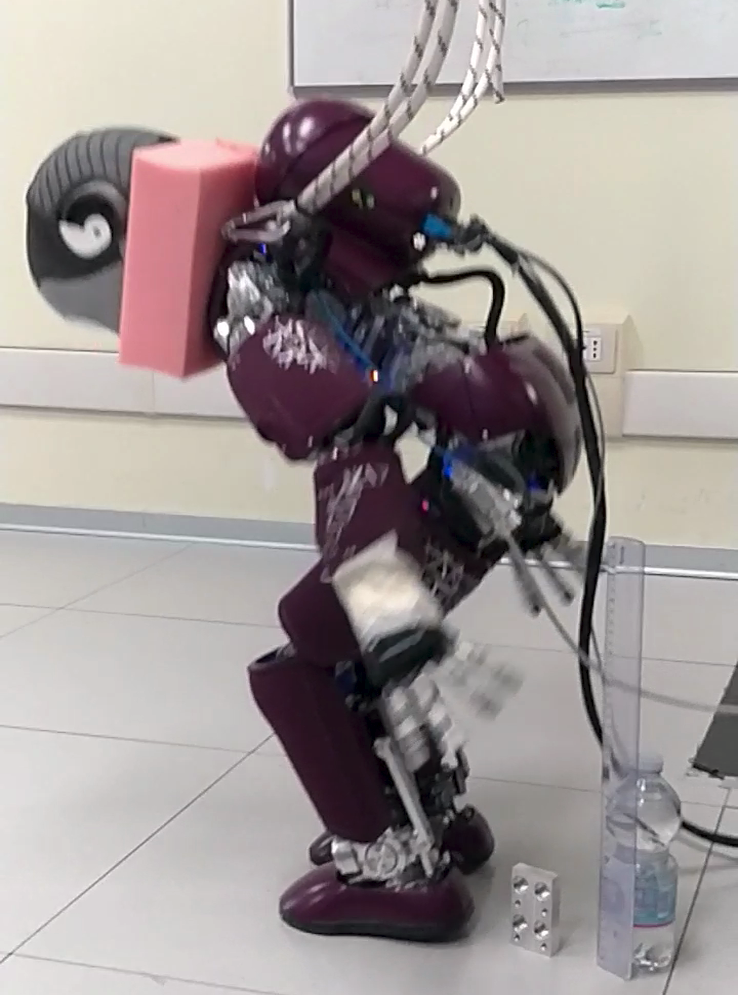}
            \caption[]%
            {{First impact}}    
            \label{fig:mean and std of net24}
        \end{subfigure}
        \begin{subfigure}[b]{0.4\columnwidth}  
            \centering 
            \includegraphics[width=\textwidth]{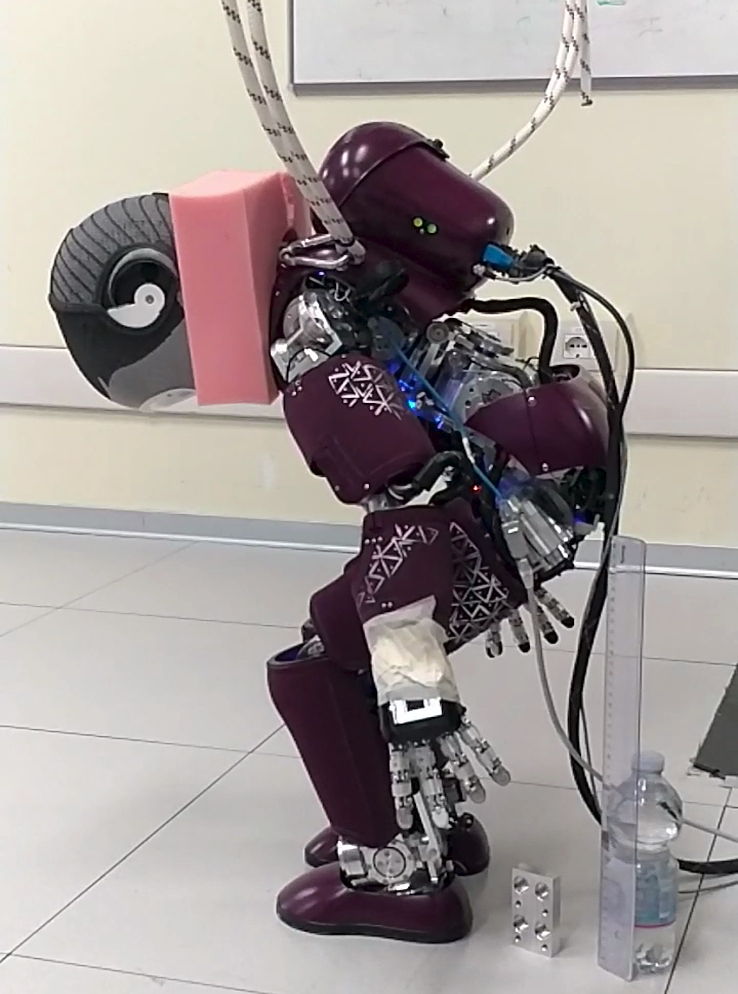}
            \caption[]%
            {{Final Configuration}}    
            \label{fig:mean and std of net24}
        \end{subfigure}
        \caption[ Velocity Acceleration profiles ]
        {\small Snapshots  extracted  from  the  accompanying  video. The robot is jumping using the velocity approach.} 
        \label{fig:CoMSpeed}
\end{figure*}

\subsection{Constraint on Angular Momentum}
\begin{figure}[H]
    \centering
    \includegraphics[width=\columnwidth]{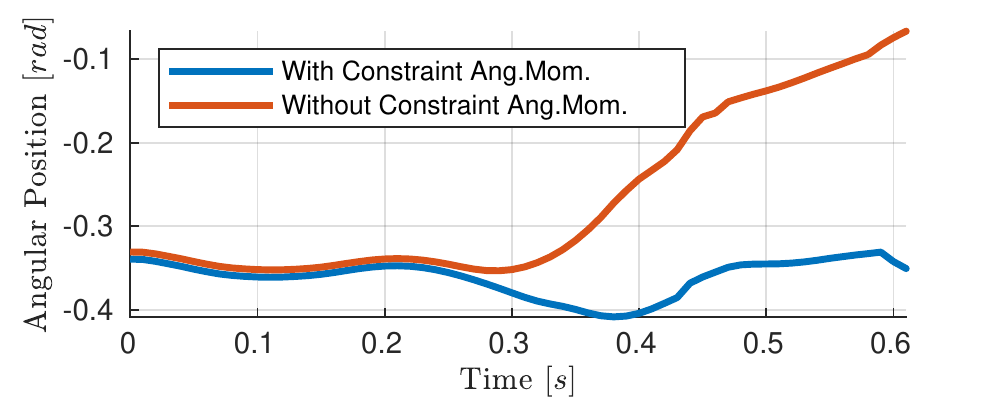}
    \caption[Base Pitch Angle]{\small Comparison between the base pitch orientation of two different simulation, the second is characterized by the absence of the constraint on the angular momentum. In both the desired jumping height is $4 [cm]$ and the variation of the position of the center of mass in the launching phase equal to $11 [cm]$.} 
        \label{fig:BASEpitchPos}
\end{figure}

In section \ref{sec:controller} the control algorithm has been described, in particular the constraints \ref{eq: AngularMomentumConstraint} and \ref{eq: AngularMomentumConstraint Torque} require a zero angular momentum. Other authors have used this approach like \cite{ResolvedMomentumControl,Nagasaka,Dai2014WholebodyMP} in jumping but also in order locomotion movements. To stress the importance of having a null angular momentum we've tried to perform jumps in simulation and experiments without controlling the angular momentum.
As could be seen in the attached video the results confirm the importance of controlling and requiring a null angular momentum. In case we neglect it, the robot has behaviour that generates a rotation of the robot while it is in the air.
In Fig.\ref{fig:BASEpitchPos} is shown a comparison between the pitch orientation of the base in two simulations, the first one with the constraint \ref{eq: AngularMomentumConstraint}, the second one without.

\section{CONCLUSIONS} \label{sec:conclusions}
In this work we presented the development of two different jumping controllers for humanoid robots.
We validated the effectiveness of the approaches both in simulation and on a real platform.
The latter, in particular has been really challenging because was the first time that the iCub robot was used for a demanding task such as jumping.
The execution of jumps on the robot required a low level improvement of the robot hardware and software.
Nevertheless, we struggled to overcame the hardware faults due to excessive peak currents, and the low level torque control, that has shown very good performances with balancing \cite{8246884} and interaction tasks, needs to be improved.
The importance of controlling and having a null angular momentum is pointed out through simulations and experiments.

Future work is planned to address the limitations of the real robot and to improve the performances of the controller. More in details we are planning to work on the following tasks:
\begin{itemize}
    \item design a QP-based controller for the flight and landing phases to guarantee better stability and compliance;
    \item integrate an IMU on the root link to better control the flight phase;
    \item use an MPC-based trajectory generator for the CoM, taking into account the joint limitations;
    \item improve the robot low level torque control by testing low level current control.
\end{itemize}

\bibliographystyle{IEEEtran}
\bibliography{Bibliografia}

\begin{thebibliography}{10}
\providecommand{\url}[1]{#1}
\csname url@samestyle\endcsname
\providecommand{\newblock}{\relax}
\providecommand{\bibinfo}[2]{#2}
\providecommand{\BIBentrySTDinterwordspacing}{\spaceskip=0pt\relax}
\providecommand{\BIBentryALTinterwordstretchfactor}{4}
\providecommand{\BIBentryALTinterwordspacing}{\spaceskip=\fontdimen2\font plus
\BIBentryALTinterwordstretchfactor\fontdimen3\font minus
  \fontdimen4\font\relax}
\providecommand{\BIBforeignlanguage}[2]{{%
\expandafter\ifx\csname l@#1\endcsname\relax
\typeout{** WARNING: IEEEtran.bst: No hyphenation pattern has been}%
\typeout{** loaded for the language `#1'. Using the pattern for}%
\typeout{** the default language instead.}%
\else
\language=\csname l@#1\endcsname
\fi
#2}}
\providecommand{\BIBdecl}{\relax}
\BIBdecl

\bibitem{Raibert:1986:LRB:6152}
M.~H. Raibert, \emph{Legged Robots That Balance}.\hskip 1em plus 0.5em minus
  0.4em\relax Cambridge, MA, USA: Massachusetts Institute of Technology, 1986.

\bibitem{Niiyama2007}
R.~Niiyama, A.~Nagakubo, and Y.~Kuniyoshi, ``{Mowgli: A bipedal jumping and
  landing robot with an artificial musculoskeletal system},'' \emph{Proceedings
  - IEEE International Conference on Robotics and Automation}, no. April, pp.
  2546--2551, 2007.

\bibitem{AtlasBFLP}
B.~Dynamics, ``What's new, atlas?''
  \url{https://www.youtube.com/watch?v=fRj34o4hN4I}, 2017.

\bibitem{Dai2014WholebodyMP}
H.~Dai, A.~Valenzuela, and R.~Tedrake, ``Whole-body motion planning with simple
  dynamics and full kinematics,'' 2014.

\bibitem{asimoVideo}
Honda, ``Honda's all-new asimo running, jumping,''
  \url{https://www.youtube.com/watch?v=fRj34o4hN4I}, 2011.

\bibitem{Nagasaka}
K.~{Nagasaka}, Y.~{Kuroki}, S.~{Suzuki}, Y.~{Itoh}, and J.~{Yamaguchi},
  ``Integrated motion control for walking, jumping and running on a small
  bipedal entertainment robot,'' in \emph{IEEE International Conference on
  Robotics and Automation, 2004. Proceedings. ICRA '04. 2004}, vol.~4, April
  2004, pp. 3189--3194 Vol.4.

\bibitem{Sakka2005}
S.~Sakka and K.~Yokoi, ``{Humanoid vertical jumping based on force feedback and
  inertial forces optimization},'' \emph{Proceedings - IEEE International
  Conference on Robotics andSakka, S., {\&} Yokoi, K. (2005). Humanoid vertical
  jumping based on force feedback and inertial forces optimization. Proceedings
  - IEEE International Conference on Robotics and Automation, 2005(Apri}, vol.
  2005, no. April, pp. 3752--3757, 2005.

\bibitem{EulerPoicare}
J.~Marsden and T.~Ratiu, \emph{Introduction to Mechanics and Symmetry: A Basic
  Exposition of Classical Mechanical Systems}.\hskip 1em plus 0.5em minus
  0.4em\relax Springer, 1999.

\bibitem{Linthorne2001}
N.~P. Linthorne, ``{Analysis of standing vertical jumps using a force
  platform},'' \emph{American Journal of Physics}, vol.~69, no.~11, pp.
  1198--1204, 2001.

\bibitem{Pattacini2010}
U.~Pattacini, F.~Nori, L.~Natale, G.~Metta, and G.~Sandini, ``An experimental
  evaluation of a novel minimum-jerk cartesian controller for humanoid
  robots,'' in \emph{2010 IEEE/RSJ International Conference on Intelligent
  Robots and Systems}, 2010, pp. 1668--1674.

\bibitem{Natale2017}
L.~Natale, C.~Bartolozzi, D.~Pucci, A.~Wykowska, and G.~Metta, ``{iCub : The
  not-yet-finished story of building a robot child},'' vol. 1026, no. December,
  pp. 2--4, 2017.

\bibitem{Parmiggiani2012}
A.~Parmiggiani, G.~Metta, and N.~Tsagarakis, ``{The mechatronic design of the
  new legs of the iCub robot},'' \emph{IEEE-RAS International Conference on
  Humanoid Robots}, pp. 481--486, 2012.

\bibitem{Ferreau2014}
H.~Ferreau, C.~Kirches, A.~Potschka, H.~Bock, and M.~Diehl, ``{qpOASES}: A
  parametric active-set algorithm for quadratic programming,''
  \emph{Mathematical Programming Computation}, vol.~6, no.~4, pp. 327--363,
  2014.

\bibitem{RomanoWBI17Journal}
F.~Romano, S.~Traversaro, D.~Pucci, and F.~Nori, ``A whole-body software
  abstraction layer for control design of free-floating mechanical systems,''
  \emph{Journal of Software Engineering for Robotics}, 2017.

\bibitem{ResolvedMomentumControl}
S.~Kajita, F.~Kanehiro, K.~Kaneko, K.~Fujiwara, K.~Harada, K.~Yokoi, and
  H.~Hirukawa, ``Resolved momentum control: Humanoid motion planning based on
  the linear and angular momentum,'' vol.~2, 11 2003, pp. 1644 -- 1650 vol.2.

\bibitem{8246884}
G.~Nava, D.~Pucci, N.~Guedelha, S.~Traversaro, F.~Romano, S.~Dafarra, and
  F.~Nori, ``Modeling and control of humanoid robots in dynamic environments:
  Icub balancing on a seesaw,'' in \emph{2017 IEEE-RAS 17th International
  Conference on Humanoid Robotics (Humanoids)}, Nov 2017, pp. 263--270.

\end{thebibliography}

\end{document}